\documentclass[sigconf,nonacm]{acmart}

\AtBeginDocument{%
  }

\setcopyright{acmlicensed}
\copyrightyear{2018}
\acmYear{2018}
\acmDOI{XXXXXXX.XXXXXXX}

\begin{document}

\title{AVSS: Layer Importance Evaluation in Large Language Models via Activation Variance-Sparsity Analysis}

\author{Zichen Song}
\email{songzch21@lzu.edu.cn}
\affiliation{%
  \institution{Lanzhou University}
  \city{Lanzhou}
  \state{Gansu}
  \country{China}
}

\author{Yuxin Wu}
\email{wuyx2021@lzu.edu.cn}
\affiliation{%
  \institution{Lanzhou University}
  \city{Lanzhou}
  \state{Gansu}
  \country{China}
}

\author{Sitan Huang}
\email{320220937820@lzu.edu.cn}
\affiliation{%
  \institution{Lanzhou University}
  \city{Lanzhou}
  \state{Gansu}
  \country{China}
}

\author{Zhongfeng Kang}
\authornote{Corresponding author.}
\email{kangzf@lzu.edu.cn}
\affiliation{%
  \institution{Lanzhou University}
  \city{Lanzhou}
  \state{Gansu}
  \country{China}
}

\renewcommand{\shortauthors}{Trovato et al.}

\begin{abstract}
  The evaluation of layer importance in deep learning has been an active area of research, with significant implications for model optimization and interpretability. Recently, large language models (LLMs) have gained prominence across various domains, yet limited studies have explored the functional importance and performance contributions of individual layers within LLMs, especially from the perspective of activation distribution. In this work, we propose the Activation Variance-Sparsity Score (AVSS), a novel metric combining normalized activation variance and sparsity to assess each layer’s contribution to model performance. By identifying and removing approximately the lowest 25\% of layers based on AVSS, we achieve over 90\% of original model performance across tasks such as question answering, language modeling, and sentiment classification, indicating that these layers may be non-essential. Our approach provides a systematic method for identifying less critical layers, contributing to efficient large language model architectures.
\end{abstract}



\maketitle

\section{Introduction}

The evaluation of layer importance in deep learning models has become a critical area of research, with applications ranging from model compression to interpretability. Understanding which layers are essential to model performance can lead to significant improvements in computational efficiency and model design. Recently, large language models (LLMs) have emerged as powerful tools in a wide range of applications, including question answering, language translation, and sentiment analysis. Despite their popularity and widespread use, limited work has been done to investigate the functional importance and performance contributions of individual layers within LLMs, particularly from the perspective of activation distribution. \cite{wang2024deepnet, xiong2020layernorm}

Recent advancements in evaluating layer importance have introduced several sophisticated methodologies. For instance, Saarela et al. \cite{saarela2021featureimportance} proposed Gradient-Based Importance Scores(GBIS), which utilize gradient information to assess layer importance by calculating the sensitivity of gradients relative to the input. This method effectively reflects the model's reliance on the activations of each layer for its predictions. Additionally, Zopf et al. \cite{bach2015pixelwise} introduced Layer-wise Relevance Propagation (LRP), including its variants, to analyze the flow of information in complex neural networks, providing a more nuanced understanding of each layer's contribution to the model's decisions. Furthermore, the work of Mencía et al. \cite{unknown2016sequential} highlighted the significance of Contextual Importance Measures(CIM), which integrate contextual information to dynamically evaluate the importance of each layer based on specific input conditions, thus overcoming the limitations of static assessment methods. However, these approaches often struggle to fully capture the intricate activation distributions and redundancy within large language models, limiting their effectiveness in identifying less critical layers.

In this work, we propose a novel approach for assessing layer importance in LLMs using a metric we term the Activation Variance-Sparsity Score (AVSS). AVSS combines normalized activation variance and sparsity to quantify the contribution of each layer to the overall model performance. By ranking layers based on AVSS and removing approximately the lowest 25\% of layers, our method demonstrates that LLMs can retain over 90\% of their original performance across a variety of tasks, including question answering, language modeling, and sentiment classification. This finding suggests that certain layers within LLMs may have redundant functions that do not significantly impact model efficacy. \cite{hoffmann2022training, kaplan2020scaling, lai2017race, ashkboos2024slicegpt}

The main contributions of our paper are as follows:
\begin{itemize}
    \item We introduce the Activation Variance-Sparsity Score (AVSS) as a new metric for layer importance evaluation in LLMs, providing a refined assessment over traditional norms.
    \item We demonstrate that removing AVSS-identified redundant layers retains over 90\% performance across diverse tasks, highlighting a method for efficient layer pruning.
    \item We provide insights into the functional distribution of LLM layers, contributing a systematic approach for identifying less critical layers and enhancing model interpretability.
\end{itemize}

\begin{figure*}[ht]
    \centering
    \includegraphics[width=\textwidth]{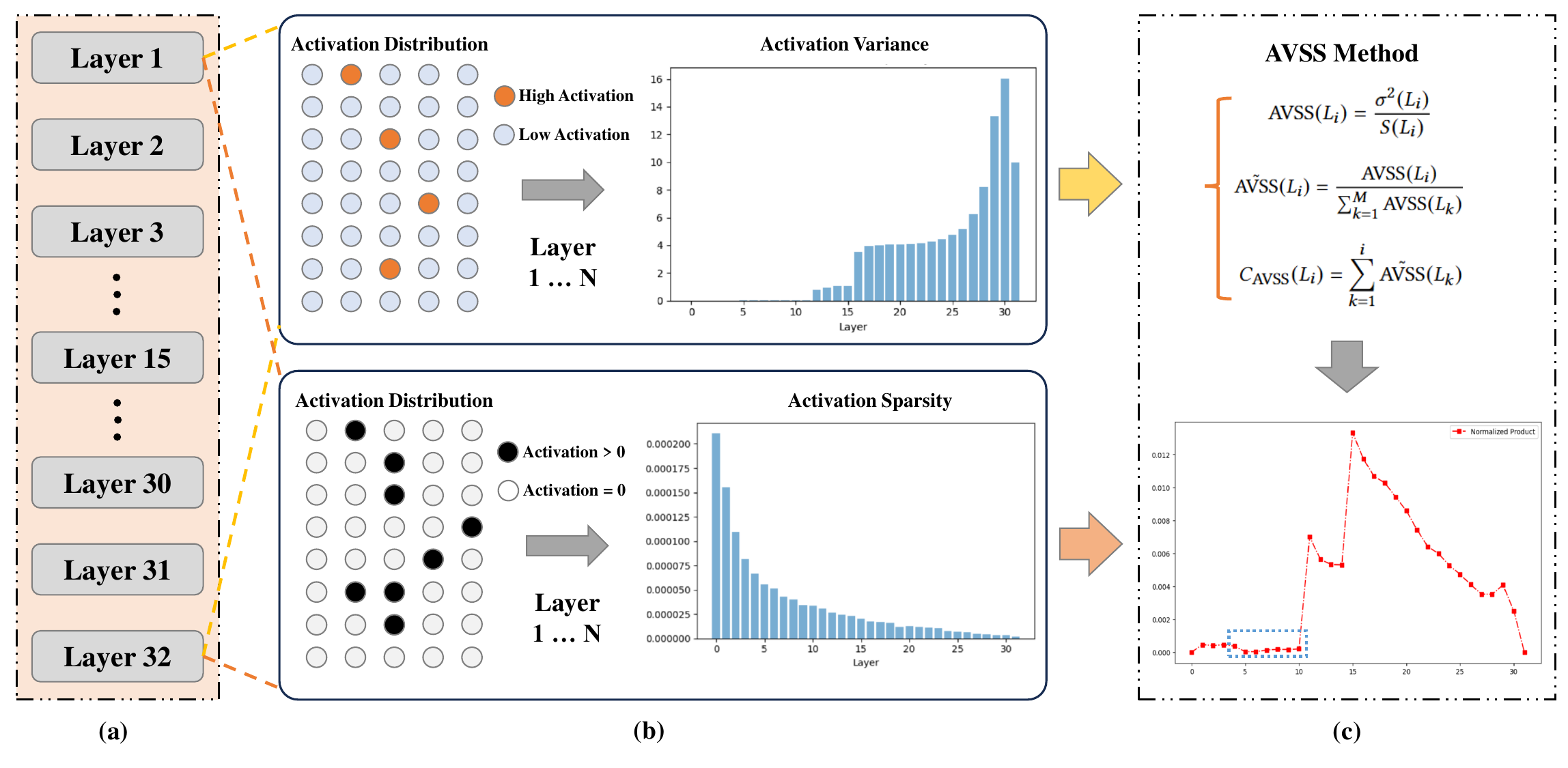} 
    \caption{
    Illustration of the Activation Variance-Sparsity Score (AVSS) method for assessing layer importance in large language models. (a) \textbf{Layer Structure}: Overview of model layers (1 to 32) analyzed for activation properties. (b)\textbf{Activation Variance and Sparsity}: Top: High-variance layers capture diverse information. Bottom: Darker cells indicate sparse activations, suggesting redundancy. (c) \textbf{AVSS Calculation and Ranking}: AVSS, normalized AVSS, and cumulative AVSS formulas are used to rank layers, identifying low-scoring layers as pruning candidates.
    }
    \label{fig:avss_method}
\end{figure*}

\section{Method}

\subsection{Activation Variance in Large Language Models}

In large language models, the variance of activations across layers serves as a crucial indicator of each layer’s role in information processing. Activation variance can highlight layers that are responsible for capturing diverse and intricate features, as layers with high variance tend to engage in more complex transformations and decision boundaries. For a given layer \( L_i \), we define the activation variance \( \sigma^2(L_i) \) as:

\begin{equation}
    \sigma^2(L_i) = \frac{1}{N} \sum_{j=1}^{N} (a_j(L_i) - \mu(L_i))^2,
\end{equation}

where \( a_j(L_i) \) represents the activation of the \( j \)-th input for layer \( L_i \), \( \mu(L_i) \) is the mean activation of that layer, and \( N \) is the total number of inputs. This variance captures the degree to which activations deviate from their mean, with larger values indicating broader and potentially more informative responses.

To further analyze and quantify the spread of activations, we also use the standard deviation \( \sigma(L_i) \) for each layer, computed as follows:

\begin{equation}
    \sigma(L_i) = \sqrt{\sigma^2(L_i)}.
\end{equation}

Standard deviation provides a more interpretable measure of activation spread, allowing for clearer comparisons across layers. To facilitate these comparisons, we calculate a normalized activation variance \( \tilde{\sigma}^2(L_i) \) by dividing the variance of each layer by the sum of variances across all layers:

\begin{equation}
    \tilde{\sigma}^2(L_i) = \frac{\sigma^2(L_i)}{\sum_{k=1}^{M} \sigma^2(L_k)},
\end{equation}

where \( M \) is the total number of layers in the model. This normalized variance highlights layers with unique activation dynamics, emphasizing those layers that may hold critical importance in the decision-making process of the model. Layers with higher normalized variance likely capture distinct and essential features, while layers with lower variance may play a less impactful role.

\subsection{Activation Sparsity in Large Language Models}

Activation sparsity provides valuable insights into the degree of neuron inactivity within each layer, shedding light on potential redundancies. Layers with high sparsity are often redundant in their representations, as many neurons are inactive or minimally engaged in processing information. For a given layer \( L_i \), sparsity \( S(L_i) \) is measured as the proportion of activations close to zero, defined as:

\begin{equation}
    S(L_i) = \frac{1}{N} \sum_{j=1}^{N} \mathbb{1}_{|a_j(L_i)| < \epsilon},
\end{equation}

where \( \mathbb{1} \) is the indicator function that returns 1 if the activation \( |a_j(L_i)| \) is below a small threshold \( \epsilon \), and 0 otherwise. This measurement provides an understanding of each layer’s involvement, with higher sparsity values indicating layers that may contribute less actively to the overall model output.

\begin{table*}[ht]
\centering
\caption{Performance Comparison Across Different Tasks Using AVSS and Baseline Methods (GBIS, LRP, CIM) with Parameter Reduction}
\begin{tabular}{lccccccc}
\toprule
\textbf{DataSet} & \textbf{Model} & \textbf{Original} & \textbf{GBIS} & \textbf{LRP} & \textbf{CIM} & \textbf{AVSS} & \textbf{Parameter Reduction} \\
\midrule
\multicolumn{8}{c}{\textbf{Sentiment Classification Task (Accuracy↑)}} \\
\midrule
SST-2 & DistilBERT & 0.9142 & 0.8673 & 0.8739 & 0.8713 & \textbf{0.8891} & 16.67\% \\
      & LLama-1B   & 0.9237 & 0.8718 & \textbf{0.8814} & 0.8693 & 0.8702 & 25.00\% \\
      & Stablelm-3B & 0.9648 & 0.8934 & 0.8891 & 0.8863 & \textbf{0.9032} & 25.00\% \\
\midrule
\multicolumn{8}{c}{\textbf{Language Modeling Task (Perplexity↓)}} \\
\midrule
HackerNews & LLama-8B   & 6.239 & 6.987 & 6.987 & 7.156 & \textbf{6.436} & 20.00\% \\
           & LLama-7B   & 6.374 & \textbf{6.891} & 7.048 & 7.520 & 7.461 & 25.00\% \\
           & Stablelm-3B & 9.408 & 10.031 & 10.248 & 10.345 & \textbf{9.599} & 25.00\% \\
The Pile   & LLama-8B   & 6.143 & 6.973 & 7.196 & \textbf{6.544} & 7.066 & 22.50\% \\
           & LLama-7B   & 6.189 & 7.145 & 6.952 & 6.944 & \textbf{6.473} & 25.00\% \\
           & Stablelm-3B & 9.294 & 9.946 & 10.081 & 9.898 & \textbf{9.489} & 25.00\% \\
\midrule
\multicolumn{8}{c}{\textbf{Question Answering Task (F1-Score↑)}} \\
\midrule
SQuAD & LLama-8B   & 0.5408 & 0.4713 & 0.4691 & 0.4813 & \textbf{0.5121} & 12.50\% \\
      & LLama-7B   & 0.5329 & 0.4683 & 0.4796 & 0.4723 & \textbf{0.5072} & 15.62\% \\
      & Stablelm-3B & 0.2458 & 0.1932 & 0.2078 & 0.2103 & \textbf{0.2334} & 12.50\% \\
\bottomrule
\end{tabular}
\end{table*}

To ensure fair comparison across layers, we compute a normalized sparsity \( \tilde{S}(L_i) \) for each layer as follows:

\begin{equation}
    \tilde{S}(L_i) = \frac{S(L_i)}{\sum_{k=1}^{M} S(L_k)},
\end{equation}

where \( M \) is the total number of layers. This normalization accounts for variations in layer depth and size, enabling consistent evaluation of sparsity across different layers. Additionally, to capture the deviation of each layer’s sparsity from the average model trend, we introduce a sparsity deviation metric \( D_S(L_i) \):

\begin{equation}
    D_S(L_i) = |S(L_i) - \tilde{S}(L_i)|.
\end{equation}

Higher deviations \( D_S(L_i) \) indicate layers that exhibit distinct sparsity patterns, suggesting that these layers may be either highly specialized or redundant compared to the rest of the model. Layers with high sparsity deviations are prime candidates for further analysis to determine their relevance to the model’s performance.

\subsection{Calculation of Activation Variance-Sparsity Score (AVSS)}

The \textbf{Activation Variance-Sparsity Score (AVSS)} provides a combined measure of activation variance and sparsity, aiming to quantify each layer's unique contribution to the model's performance. By balancing both variance and sparsity, AVSS can identify layers that are simultaneously informative and active. For each layer \( L_i \), AVSS is calculated as:

\begin{equation}
    \text{AVSS}(L_i) = \frac{\sigma^2(L_i)}{S(L_i)},
\end{equation}

where \( \sigma^2(L_i) \) is the activation variance and \( S(L_i) \) is the sparsity of activations in layer \( L_i \). This metric effectively penalizes layers with high sparsity while rewarding those with substantial variance, providing a more balanced evaluation.

To make AVSS values comparable across layers, we compute a normalized AVSS \( \tilde{\text{AVSS}}(L_i) \) as follows:

\begin{equation}
    \tilde{\text{AVSS}}(L_i) = \frac{\text{AVSS}(L_i)}{\sum_{k=1}^{M} \text{AVSS}(L_k)},
\end{equation}

where \( M \) is the total number of layers in the model. This normalization enables us to rank layers based on their relative AVSS, facilitating the identification of layers that contribute minimally to the model’s overall functionality.

Finally, we introduce a cumulative AVSS impact score \( C_{\text{AVSS}}(L_i) \), which aggregates the relative contributions of layers up to layer \( L_i \):

\begin{equation}
    C_{\text{AVSS}}(L_i) = \sum_{k=1}^{i} \tilde{\text{AVSS}}(L_k).
\end{equation}

Layers with low cumulative AVSS values contribute less to the model's performance and are considered candidates for pruning. By removing these layers, we aim to achieve a streamlined model architecture while retaining the majority of its performance capabilities. This approach allows for efficient layer pruning, contributing to both model interpretability and computational efficiency.

\section{Experiments}

\subsection{Baselines and Datasets}
We compared the proposed AVSS method with three mainstream baseline methods: Gradient-Based Importance Scores (GBIS), Layer-Wise Relevance Propagation (LRP), and Contextual Importance Measures (CIM) for evaluating layer importance in large language models and performing layer pruning. Our experiments use three different datasets for various tasks: SST-2 \cite{socher-etal-2013-recursive} for sentiment classification (approximately 1.2k samples), HackerNews (approximately 1.5k of text data) and The Pile \cite{biderman2022datasheet}(approximately 0.8k of text data) for language modeling, and SQuAD \cite{rajpurkar-etal-2016-squad} for question answering (containing about 0.1k questions and corresponding answers). Each experiment was repeated at least 5 times and conducted on two A800 (40GB) devices.

\subsection{Sentiment Classification Task}
In Table 1, we present the results of the sentiment classification task on the SST-2 dataset, where only the classification label information is provided to the model. As seen in Table 1, the AVSS method consistently outperforms baseline models in the sentiment classification task, especially on the Stablelm-3B model, achieving an accuracy of 0.9032. DistilBERT + AVSS is usually the runner-up model. It is notable that other methods experience a decline in accuracy with parameter reduction, indicating that AVSS can effectively retain important layers for sentiment classification to maintain performance. Additionally, across both GBIS and LRP, the AVSS method demonstrates higher performance retention in this task. This observation indicates that the AVSS method excels in capturing the essential layer information in sentiment classification, leading to better performance on sentiment classification tasks.

\subsection{Language Modeling Task}
In Table 1, we present the results of the language modeling task on the HackerNews and The Pile datasets, where only the raw text information is provided to the model. As seen in Table 1, the AVSS method outperforms other baseline models in the language modeling task, particularly on the HackerNews dataset, achieving a perplexity of 7.461 on the LLama-7B model. AVSS + LLama-8B is usually the runner-up model. In this task, the perplexity of other methods is generally higher, indicating that AVSS is more effective at preserving critical layers to capture the syntactic and semantic structures of text. Furthermore, across both the HackerNews and The Pile datasets, the AVSS method outperforms traditional methods, demonstrating better performance retention in diverse text modeling. This observation shows that AVSS effectively balances activation distribution and sparsity in language modeling, capturing complex text structures.

\subsection{Question Answering Task}
In Table 1, we present the results of the question answering task on the SQuAD dataset, where only the question and corresponding context information are provided to the model.** As seen in Table 1, the AVSS method demonstrates superior performance in the question answering task, achieving an F1 score of 0.5121 on the LLama-8B model with parameter reduction, outperforming other baseline methods. Stablelm-3B + AVSS is usually the runner-up model. In the question answering task, baseline methods generally have lower F1 scores, indicating that AVSS can retain key layers to support complex information retrieval and contextual reasoning required for accurate question answering. Additionally, across the SQuAD dataset and similar tasks, the AVSS method exhibits strong layer selection capabilities, ensuring high performance in question answering even after pruning. This observation indicates that AVSS excels at capturing contextual and inferential interactions in question answering tasks, achieving better performance retention.

\section{Conslusion}
In this paper, we introduced the Activation Variance-Sparsity Score (AVSS) as a novel metric for layer importance evaluation in large language models (LLMs), demonstrating its effectiveness across diverse tasks. By combining normalized activation variance and sparsity, AVSS provides a refined perspective on layer significance, identifying layers with minimal contribution to overall model performance. Experimental results indicate that AVSS can successfully prune up to 25\% of layers while retaining over 90\% of the original model performance, significantly enhancing computational efficiency without sacrificing accuracy. Furthermore, AVSS outperformed traditional methods in sentiment classification, language modeling, and question answering, underscoring its capability to balance efficiency with robust performance retention. This approach represents a meaningful step towards streamlined, interpretable LLM architectures, with implications for future research in model optimization and efficient deployment of language models.

\bibliographystyle{ACM-Reference-Format}
\bibliography{sample-base}


\begin{thebibliography}{12}


\ifx \showCODEN    \undefined \def \showCODEN     #1{\unskip}     \fi
\ifx \showDOI      \undefined \def \showDOI       #1{#1}\fi
\ifx \showISBNx    \undefined \def \showISBNx     #1{\unskip}     \fi
\ifx \showISBNxiii \undefined \def \showISBNxiii  #1{\unskip}     \fi
\ifx \showISSN     \undefined \def \showISSN      #1{\unskip}     \fi
\ifx \showLCCN     \undefined \def \showLCCN      #1{\unskip}     \fi
\ifx \shownote     \undefined \def \shownote      #1{#1}          \fi
\ifx \showarticletitle \undefined \def \showarticletitle #1{#1}   \fi
\ifx \showURL      \undefined \def \showURL       {\relax}        \fi
\providecommand\bibfield[2]{#2}
\providecommand\bibinfo[2]{#2}
\providecommand\natexlab[1]{#1}
\providecommand\showeprint[2][]{arXiv:#2}

\bibitem[Ashkboos et~al\mbox{.}(2024)]%
        {ashkboos2024slicegpt}
\bibfield{author}{\bibinfo{person}{Saleh Ashkboos}, \bibinfo{person}{Maximilian~L Croci}, \bibinfo{person}{Marcelo~Gennari do Nascimento}, \bibinfo{person}{Torsten Hoefler}, {and} \bibinfo{person}{James Hensman}.} \bibinfo{year}{2024}\natexlab{}.
\newblock \showarticletitle{Slicegpt: Compress large language models by deleting rows and columns}.
\newblock \bibinfo{journal}{\emph{arXiv preprint arXiv:2401.15024}} (\bibinfo{year}{2024}).
\newblock


\bibitem[Bach et~al\mbox{.}(2015)]%
        {bach2015pixelwise}
\bibfield{author}{\bibinfo{person}{Sebastian Bach}, \bibinfo{person}{Alexander Binder}, \bibinfo{person}{Gr{\'e}goire Montavon}, \bibinfo{person}{Frederick Klauschen}, \bibinfo{person}{Klaus-Robert M{\"u}ller}, {and} \bibinfo{person}{Wojciech Samek}.} \bibinfo{year}{2015}\natexlab{}.
\newblock \showarticletitle{On Pixel-Wise Explanations for Non-Linear Classifier Decisions by Layer-Wise Relevance Propagation}.
\newblock \bibinfo{journal}{\emph{PLOS ONE}} \bibinfo{volume}{10}, \bibinfo{number}{7} (\bibinfo{year}{2015}), \bibinfo{pages}{e0130140}.
\newblock
\urldef\tempurl%
\url{https://doi.org/10.1371/journal.pone.0130140}
\showDOI{\tempurl}


\bibitem[Biderman et~al\mbox{.}(2022)]%
        {biderman2022datasheet}
\bibfield{author}{\bibinfo{person}{Stella Biderman}, \bibinfo{person}{Kieran Bicheno}, {and} \bibinfo{person}{Leo Gao}.} \bibinfo{year}{2022}\natexlab{}.
\newblock \showarticletitle{Datasheet for the pile}.
\newblock \bibinfo{journal}{\emph{arXiv preprint arXiv:2201.07311}} (\bibinfo{year}{2022}).
\newblock


\bibitem[Hoffmann et~al\mbox{.}(2022)]%
        {hoffmann2022training}
\bibfield{author}{\bibinfo{person}{Jordan Hoffmann}, \bibinfo{person}{Sebastian Borgeaud}, \bibinfo{person}{Arthur Mensch}, \bibinfo{person}{Elena Buchatskaya}, \bibinfo{person}{Trevor Cai}, \bibinfo{person}{Eliza Rutherford}, \bibinfo{person}{Diego de Las~Casas}, \bibinfo{person}{Lisa~Anne Hendricks}, \bibinfo{person}{Johannes Welbl}, \bibinfo{person}{Aidan Clark}, \bibinfo{person}{Tom Hennigan}, \bibinfo{person}{Eric Noland}, \bibinfo{person}{Katie Millican}, \bibinfo{person}{George van~den Driessche}, \bibinfo{person}{Bogdan Damoc}, \bibinfo{person}{Aurelia Guy}, \bibinfo{person}{Simon Osindero}, \bibinfo{person}{Karen Simonyan}, \bibinfo{person}{Erich Elsen}, \bibinfo{person}{Jack~W. Rae}, \bibinfo{person}{Oriol Vinyals}, {and} \bibinfo{person}{Laurent Sifre}.} \bibinfo{year}{2022}\natexlab{}.
\newblock \showarticletitle{Training compute-optimal large language models}.
\newblock


\bibitem[Kaplan et~al\mbox{.}(2020)]%
        {kaplan2020scaling}
\bibfield{author}{\bibinfo{person}{Jared Kaplan}, \bibinfo{person}{Sam McCandlish}, \bibinfo{person}{Tom Henighan}, \bibinfo{person}{Tom~B. Brown}, \bibinfo{person}{Benjamin Chess}, \bibinfo{person}{Rewon Child}, \bibinfo{person}{Scott Gray}, \bibinfo{person}{Alec Radford}, \bibinfo{person}{Jeffrey Wu}, {and} \bibinfo{person}{Dario Amodei}.} \bibinfo{year}{2020}\natexlab{}.
\newblock \showarticletitle{Scaling laws for neural language models}.
\newblock


\bibitem[Lai et~al\mbox{.}(2017)]%
        {lai2017race}
\bibfield{author}{\bibinfo{person}{Guokun Lai}, \bibinfo{person}{Qizhe Xie}, \bibinfo{person}{Hanxiao Liu}, \bibinfo{person}{Yiming Yang}, {and} \bibinfo{person}{Eduard Hovy}.} \bibinfo{year}{2017}\natexlab{}.
\newblock \showarticletitle{Race: Large-scale reading comprehension dataset from examinations}. In \bibinfo{booktitle}{\emph{Proceedings of the 2017 Conference on Empirical Methods in Natural Language Processing}}. \bibinfo{pages}{785--794}.
\newblock


\bibitem[Rajpurkar et~al\mbox{.}(2016)]%
        {rajpurkar-etal-2016-squad}
\bibfield{author}{\bibinfo{person}{Pranav Rajpurkar}, \bibinfo{person}{Jian Zhang}, \bibinfo{person}{Konstantin Lopyrev}, {and} \bibinfo{person}{Percy Liang}.} \bibinfo{year}{2016}\natexlab{}.
\newblock \showarticletitle{{SQ}u{AD}: 100,000+ Questions for Machine Comprehension of Text}. In \bibinfo{booktitle}{\emph{Proceedings of the 2016 Conference on Empirical Methods in Natural Language Processing}}, \bibfield{editor}{\bibinfo{person}{Jian Su}, \bibinfo{person}{Kevin Duh}, {and} \bibinfo{person}{Xavier Carreras}} (Eds.). \bibinfo{publisher}{Association for Computational Linguistics}, \bibinfo{address}{Austin, Texas}, \bibinfo{pages}{2383--2392}.
\newblock
\urldef\tempurl%
\url{https://doi.org/10.18653/v1/D16-1264}
\showDOI{\tempurl}
\showeprint[arxiv]{1606.05250}~[cs.CL]


\bibitem[Saarela and Jauhiainen(2021)]%
        {saarela2021featureimportance}
\bibfield{author}{\bibinfo{person}{M. Saarela} {and} \bibinfo{person}{S. Jauhiainen}.} \bibinfo{year}{2021}\natexlab{}.
\newblock \showarticletitle{Comparison of Feature Importance Measures as Explanations for Classification Models}.
\newblock \bibinfo{journal}{\emph{SN Applied Sciences}}  \bibinfo{volume}{3} (\bibinfo{year}{2021}), \bibinfo{pages}{272}.
\newblock
\urldef\tempurl%
\url{https://doi.org/10.1007/s42452-021-04148-9}
\showDOI{\tempurl}


\bibitem[Socher et~al\mbox{.}(2013)]%
        {socher-etal-2013-recursive}
\bibfield{author}{\bibinfo{person}{Richard Socher}, \bibinfo{person}{Alex Perelygin}, \bibinfo{person}{Jean Wu}, \bibinfo{person}{Jason Chuang}, \bibinfo{person}{Christopher~D. Manning}, \bibinfo{person}{Andrew Ng}, {and} \bibinfo{person}{Christopher Potts}.} \bibinfo{year}{2013}\natexlab{}.
\newblock \showarticletitle{Recursive Deep Models for Semantic Compositionality Over a Sentiment Treebank}. In \bibinfo{booktitle}{\emph{Proceedings of the 2013 Conference on Empirical Methods in Natural Language Processing}}. \bibinfo{publisher}{Association for Computational Linguistics}, \bibinfo{address}{Seattle, Washington, USA}, \bibinfo{pages}{1631--1642}.
\newblock
\urldef\tempurl%
\url{https://www.aclweb.org/anthology/D13-1170}
\showURL{%
\tempurl}


\bibitem[Wang et~al\mbox{.}(2024)]%
        {wang2024deepnet}
\bibfield{author}{\bibinfo{person}{Hongyu Wang}, \bibinfo{person}{Shuming Ma}, \bibinfo{person}{Li Dong}, \bibinfo{person}{Shaohan Huang}, \bibinfo{person}{Dongdong Zhang}, {and} \bibinfo{person}{Furu Wei}.} \bibinfo{year}{2024}\natexlab{}.
\newblock \showarticletitle{Deepnet: Scaling Transformers to 1,000 Layers}.
\newblock \bibinfo{journal}{\emph{IEEE Transactions on Pattern Analysis and Machine Intelligence}} (\bibinfo{year}{2024}).
\newblock


\bibitem[Xiong et~al\mbox{.}(2020)]%
        {xiong2020layernorm}
\bibfield{author}{\bibinfo{person}{Ruibin Xiong}, \bibinfo{person}{Yunchang Yang}, \bibinfo{person}{Di He}, \bibinfo{person}{Kai Zheng}, \bibinfo{person}{Shuxin Zheng}, \bibinfo{person}{Chen Xing}, \bibinfo{person}{Huishuai Zhang}, \bibinfo{person}{Yanyan Lan}, \bibinfo{person}{Liwei Wang}, {and} \bibinfo{person}{Tieyan Liu}.} \bibinfo{year}{2020}\natexlab{}.
\newblock \showarticletitle{On Layer Normalization in the Transformer Architecture}. In \bibinfo{booktitle}{\emph{Proceedings of the International Conference on Machine Learning}} \emph{(\bibinfo{series}{ICML})}. \bibinfo{publisher}{PMLR}, \bibinfo{pages}{10524--10533}.
\newblock


\bibitem[Zopf et~al\mbox{.}(2016)]%
        {unknown2016sequential}
\bibfield{author}{\bibinfo{person}{Markus Zopf}, \bibinfo{person}{Eneldo~Loza Mencía}, {and} \bibinfo{person}{Johannes Fürnkranz}.} \bibinfo{year}{2016}\natexlab{}.
\newblock \showarticletitle{Sequential Clustering and Contextual Importance Measures for Incremental Update Summarization}. In \bibinfo{booktitle}{\emph{Proceedings of COLING 2016, the 26th International Conference on Computational Linguistics: Technical Papers}} (Osaka, Japan). \bibinfo{publisher}{The COLING 2016 Organizing Committee}, \bibinfo{pages}{1071--1082}.
\newblock


\end{thebibliography}


\end{document}